\documentclass[10pt,twocolumn,letterpaper]{article}

\usepackage{wacv}
\usepackage{times}
\usepackage{epsfig}
\usepackage{graphicx}
\usepackage{amsmath}
\usepackage{amssymb}
\usepackage{booktabs}
\usepackage{color}

\wacvfinalcopy 

\ifwacvfinal\fi
\begin{document}

\title{Efficient Object Detection in Large Images using Deep Reinforcement Learning}

\author{Burak Uzkent \\
Department of Computer Science\\
Stanford University\\
{\tt\small buzkent@cs.stanford.edu}
\and
Christopher Yeh \\
Department of Computer Science\\
Stanford University \\
{\tt\small chrisyeh@stanford.edu}
\and
Stefano Ermon \\
Department of Computer Science\\
Stanford University \\
{\tt\small ermon@cs.stanford.edu}
}

\maketitle
\ifwacvfinal\thispagestyle{empty}\fi

\begin{abstract}
    Traditionally, an object detector is applied to every part of the scene of interest, and its accuracy and computational cost increases with higher resolution images. However, in some application domains such as remote sensing, purchasing high spatial resolution images is expensive. To reduce the large computational and monetary cost associated with using high spatial resolution images, we propose a reinforcement learning agent that adaptively selects the spatial resolution of each image that is provided to the detector. In particular, we train the agent in a dual reward setting to choose low spatial resolution images to be run through a coarse level detector when the image is dominated by large objects, and high spatial resolution images to be run through a fine level detector when it is dominated by small objects. This reduces the dependency on high spatial resolution images for building a robust detector and increases run-time efficiency. We perform experiments on the xView dataset, consisting of large images, where we increase run-time efficiency by $50\%$ and use high resolution images only $30\%$ of the time while maintaining similar accuracy as a detector that uses only high resolution images.
\end{abstract}

\section{Introduction}
Deep Convolutional Neural Networks have been successfully applied to different computer vision tasks including image recognition~\cite{krizhevsky2012imagenet,he2016deep,simonyan2014very}, object detection~\cite{lin2014microsoft,ren2015faster,redmon2017yolov2}, object tracking~\cite{kristan2017visual,bertinetto2016fully,uzkent2017aerial,uzkent2018tracking,uzkent2018enkcf}. Traditionally, the CNN's use images resized to predetermined number of pixels for each dimension. For instance, for image recognition, typically we resize original images to 224$\times$224px to utilize ImageNet pre-trained weights~\cite{huh2016makes,hendrycks2019using}. 

On the other hand, convolutional object detectors are traditionally applied to images that are resized to dimensions less than 1,000 pixels on each side (e.g.$\sim$500px for images from the MSCOCO and PASCAL VOC2007/VOC2012 datasets \cite{liu2016ssd}, and $\sim$600px for satellite images from the xView dataset~\cite{sergievskiy2019reduced}). However, in some application domains, images may be much larger, on the order of several thousands of pixels in each dimension. These large images enable higher detection accuracy, especially for smaller objects relative to the field of view \cite{redmon2017yolov2}, but they require additional computation, time, and/or financial resources to process.

For example, in the field of self-driving vehicles or traffic monitoring, fusing together images from multiple sensors produces large images, which presents a computational challenge to maintain real-time processing~\cite{zhu2016traffic,meng2017detecting}.
Similarly, most satellite images in the xView dataset have about 1,000-10,000px on each side~\cite{lam2018xview}. Yet, in realistic remote sensing applications, these high spatial resolution satellite images (e.g.$<$1m/px) are financially costly to acquire, compared to publicly available low spatial resolution satellite images (e.g. 10m/px)~\cite{drusch2012sentinel2}. Therefore, it is desirable to build a system that will utilize the dependency on large images to reduce the costs of analyzing satellite images for computer vision tasks including image recognition~\cite{sheehan2018learning,uzkent2019learning}, object detection~\cite{lam2018xview,chen2014vehicle}, object tracking~\cite{uzkent2016real,uzkent2014feature}, and poverty mapping~\cite{sheehan2019predicting}.

Directly applying existing state-of-the-art convolutional detectors on large images not only increases processing time but also the memory required to store large feature maps~\cite{lampert2008beyond,wojek2008sliding}. The traditional \textit{sliding window} approach mitigates the memory requirement by ``sliding'' the convolutional detector over crops of the image until the full image has been processed. However, the processing time of this technique increases quadratically with respect to side-length of the image, and it may have lower accuracy because the detector is unable to ``see'' all of the image at once.

Considering the trade-off between \emph{accuracy} and various \emph{costs} associated with using high spatial resolution images, we propose an adaptive framework that only chooses high spatial resolution images when fine level information is required, and uses low spatial resolution images when the coarse level information provides sufficient information for the objects of interest. This maintains the overall detection accuracy while increasing run-time efficiency and reducing the dependency on expensive, high resolution (HR) imagery.

To describe our approach in high level, we train an agent using \emph{reinforcement learning} to explicitly maximize accuracy while minimizing the amount of HR images. It first processes a low resolution (LR) image, learning the global image context, then selectively requests certain HR patches that it deems necessary for making accurate object detections. Importantly, the agent is not given the HR image, thus limiting potential acquisition costs for the HR image to only the patches that it selects.

Experiments on the xView dataset demonstrate that compared to an object detector that exclusively uses HR images, our method maintains nearly the same accuracy but uses HR images only about 30\% of the time, in turn increasing run-time performance by about $50\%$.

\section{Related Work}


\textbf{Convolutional Object Detection}
Earlier convolutional object detectors such as R-CNN and Fast-RCNN~\cite{girshick2014rcnn,girshick2015fast} use a selective search algorithm~\cite{uijlings2013selective} to identify box proposals to run the detector on. On the other hand, Faster-RCNN~\cite{ren2017fasterrcnn} jointly learns box proposals and the classifier in an end-to-end manner~\cite{ren2017fasterrcnn}, but the two-stage design limits computational efficiency. Instead, single-shot detectors~\cite{liu2016ssd, redmon2018yolov3,lin2017focal} directly predict bounding boxes from fixed anchors on input images without relying on a region proposal network, thereby achieving real-time performance. To replace fixed anchors with sparse adaptive anchors, \cite{lu2016adaptive} propose a search method to predict anchor boxes that is likely to contain objects. Their method improves efficiency on cases with sparse object instances.

\textbf{Sliding Windows on Large Images}
Running single and two stage detectors on large images requires a large amount of memory to store large feature maps \cite{redmon2017yolov2}. To avoid that, the traditional sliding window technique divides the image into smaller (often overlapping) windows and then runs the detector on each window~\cite{harzallah2009combining}. While this approach has the same memory footprint as a detector running on a single window, the number of windows (and therefore run time) increases quadratically in the image side length. Furthermore, this technique is wasteful if the detector is run on a window that contains no objects or has objects large enough to be detected in a LR image.

\textbf{Pruning the image search space}
Various techniques have been proposed to reduce the search space, such as selectively choosing windows to evaluate based on previously observed windows~\cite{alexe2012searching, gonzalez-garcia2015active, najibi2018autofocus}. However, such systems introduce non-trivial overhead on the order of seconds per image, and they are not designed to take advantage of the global image context. Other works adopt a cascade of object detectors to narrow the search space: \cite{li2015cnncascade} starts with downsampling the sliding windows to 12$\times$12 px images to process with a shallow 12Net to easily remove negatives. They then process the faces detected by 12Net using the 24Net with 24$\times$24 px images. 
On the other hand, \cite{lalonde2018clusternet} adopts a two-stage CNN where the first stage learns a rough heat map of the moving vehicles whereas the second stage processes the chosen ROIs to localize the vehicles. However, both~\cite{li2015cnncascade, lalonde2018clusternet} assume access to the HR image corresponding to every object in the image, which may be costly to obtain, and \cite{lalonde2018clusternet} additionally uses temporally consecutive video frames of the same region, which are not available everywhere and only help locate non-stationary objects.

As in~\cite{lalonde2018clusternet}, we adopt a two-stage approach, but we train a RL agent using single LR image to choose the patches to zoom-in when the gain with fine detector on HR patch is higher than the gain with coarse detector on LR patch.

\textbf{Reinforcement Learning for Efficient Detection}
Reinforcement Learning (RL) has been recently used to (1) \emph{replace classical detectors such as SSD and Faster-RCNN}, (2) \emph{replace exhaustive box proposal techniques in two-stage detectors}, and (3) \emph{find ROIs in very large images to run a detector on}. 
Most of the methods proposed in this categories focus on learning \emph{sequential policies}. Under category (1),~\cite{caicedo2015active, mathe2016reinforcement} proposed a top-down sequential object detection models trained with Q-learning algorithm. \cite{caicedo2015active} uses an action space that deforms the bounding box by applying translation and scale factors whereas~\cite{mathe2016reinforcement} uses actions to fixate a bounding box in image space. Most of the RL methods associated with object detection fall into category (2). For example, \cite{jie2016tree} recursively divides up an image in a top-down approach where the divisions are decided by the RL agent. The box proposals returned by the agent are then passed through Fast-RCNN.
Some other studies use RL for sequentially finding box proposals to replace the first stage of two-stage detectors~\cite{gonzalez-garcia2015active,pirinen2018deep}.
Our approach, like \cite{gao2018dynamic}, falls into category (3) where a RL agent is trained to examine a down-sampled image and sequentially choose ROIs to zoom-in on. For efficiency, \cite{gao2018dynamic} directly use the detections on the full down-sampled image if the improvement gained by zooming-in to a ROI is not sufficiently high.
Unlike~\cite{jie2016tree}, they use a \emph{cost-sensitive} reward function to limit number the of steps as in ~\cite{he2012cost,karayev2014anytime}. Similarly, we learn the zoom-in policies with a \emph{cost-sensitive} reward function.
On the other hand, they downsample the initial large image by \emph{2} and focus on pedestrians represented by reasonable number of pixels. In contrast, we consider a higher downsampling ratio for the policy network on very large remote sensing images (e.g. $>$1000px on each side) in which objects are represented by drastically \emph{small number of pixels}. Additionally, similarly to our study, their zoom-in ROIs can contain \emph{multiple instances} of objects whereas~\cite{gonzalez-garcia2015active,mathe2016reinforcement,pirinen2018deep,jie2016tree} search for the boxes surrounding a \emph{single object}.
Finally, all of the previous methods proposed for efficient detection learn sequential policies whereas ours chooses the zoom-in ROIs in \emph{single forward pass}. This can be highly beneficial to \emph{parallelize} detection on zoom-in ROIs.



\section{Proposed Formulation}
\begin{figure}[!h]
\centering
\includegraphics[width=\linewidth]{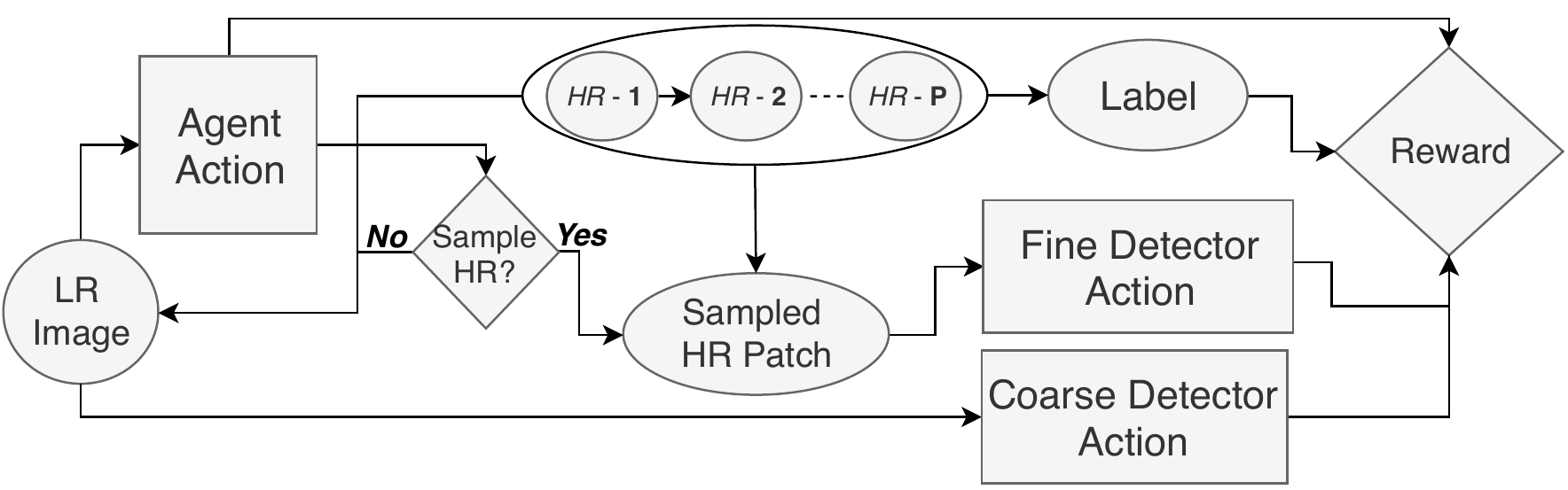}
\caption{Proposed Bayesian Decision \emph{influence} diagram. At training time, HR images are downsampled to LR images which the agent uses for sampling actions. Detector (coarse/fine) then uses LR/HR image to output bounding boxes which are used to compute reward together with the labels from HR images.}
\label{fig:task_definition}
\end{figure}
We propose an efficient object detection framework consisting of two modules named \emph{coarse} and \emph{fine} level search. With coarse level search, we perform an initial search space optimization on \emph{very large} images with over 3,000 pixels in each dimension. At the end of coarse level search, we find an initial set of patches where it may be beneficial to zoom-in and acquire HR images. In fine level search, we perform further search space optimization on the patches chosen by the coarse level search module to make a final decision about which subpatches to acquire HR images for. Both coarse and fine level search modules are formulated similarly with a two-step episodic Markov Decision Process (MDP), as shown in our generic influence diagram in Fig.~\ref{fig:task_definition}. In the diagram, we represent random variables with a circle, actions with a square/rectangle, and utilities with a diamond.


\textbf{Coarse Level Search}
This first module of our framework, implemented as \emph{CPNet}, finds ROIs/patches to zoom into, conditioned on the low spatial resolution image of the initial large area. This is achieved by applying the proposed generic influence diagram as shown in Fig.~\ref{fig:task_definition} to coarse level search. In this direction, a large HR image $x_{H} = (x_{H}^{1}, x_{H}^{2},\dotsc, x_{H}^{P_{c}})$ is composed of equal-size non-overlapping patches, where $P_{c}$ is the number of coarse-level patches. Unlike in traditional computer vision settings, $x_{H}$ is latent, \emph{i.e.}, it is \emph{not observed by the agent}. $Y=\{Y_1, \dotsc, Y_{P_c}\}$ is an array of arrays of the (unobserved) ground truth bounding boxes associated with each patch of $x_{H}$, where $Y_{i}=\{y_{1}, \dotsc, y_{P_f}\}$. Each bounding box is represented as a tuple $y_{i}^{j}= (g_{x}, g_{y}, w, h, c)$, which is a random variable containing the centroid, width, height, and object class. The random variable $x_{L} = (x_{L}^{1}, x_{L}^{2}, \dotsc, x_{L}^{P_{c}})$ denotes the LR image of the same scene as $x_{H}$, where $x_{L}^{i}$ represents the lower spatial resolution version of $x_{H}^{i}$.

In the first step of the MDP for coarse level search, the agent observes $x_{L}$ and outputs a binary action array, $a_{c} \in \{0,1\}^{P_c}$, where $a_{c}^{i} = 1$ means that the agent would like to consider acquiring HR subpatches of the $i$-th patch $x_{H}^i$. We define the patch sampling policy model, parameterized by $\theta_{p}^{c}$, as
\begin{equation}
    \pi_{c}(a_c | x_{L};\theta_{p}^{c}) = p(a_{c}|x_{L};\theta_{p}^{c})
\label{eq:policy_cpnet}
\end{equation}
where $\pi_{c}(x_{L};\theta_{p}^{c})$ is a function mapping the observed LR image to a probability distribution over patch sampling actions $a_{c}$. The joint probability distribution over the random variables $x_{H}$, $Y$, $x_{L}$, and action $a_{c}$, can be written as
\begin{align}
    &p(x_{H}, x_{L}, Y, a_c) \nonumber \\
    &= p(x_{H})\, p(Y|x_{H})\, p(x_{L}|x_{H})\, p(a_{c}|x_{L};\theta_{p}^{c}).
\label{eq:joint_distribution_cpnet}
\end{align}

In the second step of the MDP, the agent runs the object detection policy. Conditioned on $a_c$, it observes either $x_{H}^{i}$ or $x_{L}^{i}$ and chooses an action $a_{d}=\hat{Y}_{i}$ where $\hat{Y}_{i}=\{\hat{y}_{1} \cdots \hat{y}_{P_{f}}\}$ 
and $\hat{y}_{i}^{j} = (\hat{g}_{x}, \hat{g}_{y}, \hat{w}, \hat{h}, \hat{c})$ represents a predicted bounding box for $x_{H}^{i}$ or $x_{L}^{i}$. We define the object detection policy as follows:
\begin{equation}
    \pi_{d}(a_d|x_{L}^{i};\theta_{d}^{c}) = p(a_{d}|x_{L}^{i};\theta_{d}^{c}),
\label{eq:detection_coarse}
\end{equation}
\begin{equation}
    \pi_{d}(a_d|x_{H}^{i};\theta_{d}^{f}) = p(a_{d}|x_{H}^{i};\theta_{d}^{f}),
\label{eq:detection_fine}
\end{equation}
where $\theta_{d}^{c}$ and $\theta_{d}^{f}$ represent the coarse and fine object detectors operating on $x_{L}^{i}$ and $x_{H}^{i}$.

The overall objective $J_c$ is defined to maximizing the expected utility $R_c$ given the evidence, represented by
\begin{equation}
\max_{\theta_{p}^{c}, \theta_{d}^{f}, \theta_{d}^{c}} J_c(\theta_{p}^{c}, \theta_{d}^{f}, \theta_{d}^{c}) = \mathbb{E}_p[R_c(a_{c}, a_{d}, Y)],
\label{Eq:Cost_Function}
\end{equation}
where the utility depends on $a_{c}$, $a_{d}$, and $Y$. The reward penalizes the agent for selecting a large number of HR patches (e.g., based on the norm of $a_{c}$) and includes a performance metric evaluating the accuracy of detector, $a_{d}$, given the true label $Y$ (e.g., recall, precision). We detail the reward function in the section~\ref{sect:reward_function}.

\begin{figure*}[!h]
\centering
\includegraphics[width=0.99\textwidth]{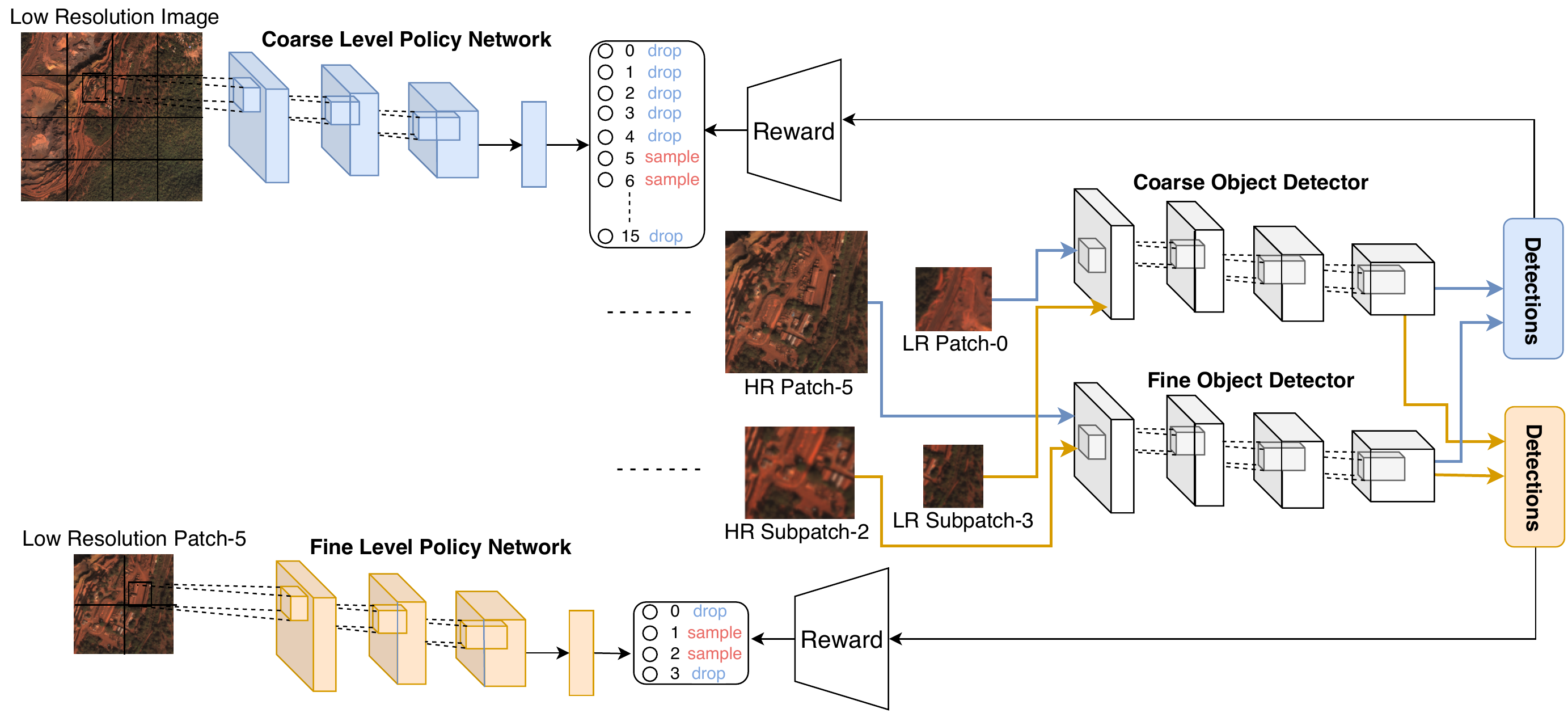}
\caption{The proposed coarse and fine level policy networks (CPNet, and FPNet) to process \emph{very large images} (e.g. $>$1000px on each side). The CPNet uses the inital large LR image to choose a set of actions representing a unique patch in the image space. The sampled patch is then either used by coarse or fine detector to estimate the expected reward. The FPNet uses a patch of an initial large LR image to choose a set of actions. Next, the coarse or fine detector is run on the sampled subpatch. We train CPNet and FPNet independently. In test time, in our cascaded approach, we first run CPNet on large image and run FPNet on patches asked to be sampled by CPNet. 
} 
\label{fig:CPNet_FPNet}
\end{figure*}
\textbf{Fine Level Search}
The CPNet alone can be satisfactory when considering images with $<$1000px on each side. However, for very large images (e.g. $>$1000px on each side), it is desirable to further optimize the search space as a \emph{patch can be represented by large number of pixels}. In such images, one way to further optimize the search space is to reduce the size of patches for CPNet, resulting in larger action space. However, training CPNet with larger action space can be unstable and take drastically longer time. For this reason, we use another policy network, called \emph{FPNet}, parameterized by $\theta_{p}^{f}$, and apply it to the random variable low spatial resolution patch $x_{L}^{i}$ sampled from the latent variable $x_{H}^{i} = (x_{h}^{1}, x_{h}^{2}, \dotsc, x_{h}^{P_{f}})$ consisting of $P_f$ overlapping fine-level subpatches. 
The task of this policy network is to choose a binary action array $a_{f} \in \{0,1\}^{P_{f}}$, where $a_f^j =1$ means that the agent would like to acquire the $j$-th HR subpatch $x_{h}^{j}$. The subpatch sampling policy model is parameterized by $\theta_{p}^{f}$ and formulated similarly to Eq~\ref{eq:policy_cpnet}.

The first step of the MDP can then be modeled similarly to the CPNet with a joint probability distribution over the random variables as
\begin{align}
    p(x_{H}^{i}, x_{L}^{i}, Y^{i}, a_f) =  p(x_{H}^{i})p(Y^{i}|x_{H}^{i}) \\ p(x_{L}^{i}|x_{H}^{i}) \nonumber
    p(a_{f}|x_{L}^{i};\theta_{p}^{f}).
\end{align}

In the second step of the MDP, the agent observes the random variables, $x_{h}^{j}$ or $x_{l}^{j}$ for $j=\{1, \dotsc, P_{f}\}$, based on $a_{f}$ and chooses an action $a_{d}=\hat{Y}_{i}$, where
$\hat{Y}_{i}=\{\hat{y}_{1}, \dotsc, \hat{y}_{P_f}\}$ and
$\hat{y}_{i}^{m} = (\hat{g}_{x}, \hat{g}_{y}, \hat{w}, \hat{h}, \hat{c})$ represents a predicted bounding box for $x_{h}^{j}$ or $x_{l}^{j}$. We then define the object detection policy similarly to Eq~\ref{eq:detection_coarse} and~\ref{eq:detection_fine}.
Finally, the overall objective function $J_f$ is  defined as maximizing the expected utility $R_f$ given the evidence, represented by
\begin{equation}
\max_{\theta_{p}^{f}, \theta_{d}^{f}, \theta_{d}^{c}} J_f(\theta_{p}^{f}, \theta_{d}^{f}, \theta_{d}^{c}) = \mathbb{E}_p[R_f(a_{f}, a_{d}, Y_{i})].
\label{Eq:Cost_Function}
\end{equation}
Fig.~\ref{fig:CPNet_FPNet} visualizes the proposed CPNet and FPNet in detail.
\section{Proposed Solution}
\subsection{Modeling the Policy Networks and Detectors}
In the previous section, we formulated the task of efficient object detection in a cascaded approach where coarse and fine level search is formulated as a two step episodic MDP. Here, we detail the action space and how the policy distributions for $a_{c}$, $a_{f}$ and $a_{d}$ are modelled for each module. To represent our discrete action space for $a_{c}$ and $a_{f}$, we divide the image space into equal size patches, resulting in $P_c$ and $P_f$ number of patches and subpatches. In this study, for the coarse and fine level search, we use $P_c=16$ and $P_f=4$ regardless of the size of the input image and leave the task of choosing variable size bounding boxes as a future work. In the first step of the two step MDP, the policy networks, $f_{p}^c$ and $f_{p}^f$, \emph{output the probabilities for all the actions at once} after observing $x_{L}$ and $x_{L}^{i}$. An alternative approach could be in the form of a framework where $a_{c,f}^{i,j}$ is conditioned on $a_{c,f}^{1:i-1,1:j-1}$. However, the proposed concept of \emph{outputting all the actions at once} provides a more efficient decision making process.

In this study, similarly to~\cite{wu2018blockdrop}, we model the action likelihood function of the policy networks, $f_{p}^c$ and $f_{p}^f$, by multiplying the probabilities of the individual HR patch/subpatch selections, represented by Bernoulli distributions as follows:
\begin{align}
  \pi_{c}(a_{c}|x_{L},\theta_{p}^{c}) &= \prod_{i=1}^{P_c} s_{c}^{i}(1- s_{c}^{i})^{(1-a_{c}^{i})}, \\
  \pi_{f}(a_{f}|x_{L}^{i},\theta_{p}^{f}) &= \prod_{j=1}^{P_f} s_{f}^{j}(1- s_{f}^{j})^{(1-a_{f}^{j})}
\label{eq:action_likelihood}
\end{align}
where $s$ is the prediction vector formulated as
\begin{align}
    s_{c} &= f_{p}^{c}(x_{L};\theta_{p}^{c}), \\
    s_{f} &= f_{p}^{f}(x_{L}^{i};\theta_{p}^{f}).
\label{eq:policy_network}
\end{align}

To get probabilistic values, $s_{c}, s_{f} \in [0,1]$, we use a sigmoid function on the final layers of CPNet and FPNet.

The next set of actions for the coarse level search, $a_{d}$, is chosen by the coarse and fine level object detectors using the LR patch $x_{L}^{i}$ or the sampled HR patch $x_{H}^{i}$. For the fine level search, the actions, $a_{d}$, are chosen using the LR subpatch $x_{l}^{i}$ or the sampled HR subpatch $x_{h}^{i}$.

\subsection{Training the Policy Networks} 
After defining the two step MDP and modeling the policy and detector networks, we detail the training procedure of the proposed efficient object detection model. The goal of training is to learn the optimal parameters of policy networks, $\theta_{p}^{c}$ and $\theta_{p}^{f}$. Because the actions are discrete, we cannot use the reparameterization trick to optimize the objective w.r.t. $\theta_{p}^{c}$ and $\theta_{p}^{f}$. To optimize the parameters $\theta_{p}^{c}$, $\theta_{p}^{f}$ of $f_{p}^c$ and $f_p^f$, we need to use model-free reinforcement learning algorithms such as Q-learning~\cite{Watkins92q-learning} and policy gradient~\cite{sutton2018reinforcement}. Policy gradient is more suitable in our scenario since the number of unique actions the policy network can choose is $2^{P}$ and increases exponentially with $P$. Finally, we use the \textit{REINFORCE} method~\cite{sutton2018reinforcement} to optimize the objective w.r.t. policy network parameters as:
\begin{align}
\nabla_{\theta_{p}^{c}}J_c &= \mathbb{E}\left[ R_c(a_{c}, a_{d}, Y)\nabla_{\theta_{p}^{c}} \log \pi_{\theta_{p}^{c}}(a_{c}|x_{L}) \right],
\label{eq:policy_gradient_coarse} \\
\nabla_{\theta_{p}^{f}}J_f &= \mathbb{E}\left[ R_f(a_{f}, a_{d}, Y_{i})\nabla_{\theta_{p}^{f}} \log \pi_{\theta_{p}^{f}}(a_{f}|x_{L}^{i}) \right].
\label{eq:policy_gradient_fine}
\end{align}

Averaging across a mini-batch via Monte-Carlo sampling produces an unbiased estimate of the expected value, but with potentially large variance. Since this can lead to an unstable training process~\cite{wu2018blockdrop}, we replace $R_c(a_{c}, a_{d}, Y)$ in Eq.~\ref{eq:policy_gradient_coarse} with the advantage function to reduce the variance:
\begin{equation}
\nabla_{\theta_{p}^c}J_c = \mathbb{E}\left[ A \sum_{i=1}^{P_c} \nabla_{\theta_{p}^c}\log(s_{c}^{i}a_{c}^{i}+(1-s_{c}^{i})(1-a_{c}^{i})) \right]
\end{equation}
where
\begin{equation}
    A(a_{c}, \hat{a}_{c}, a_{d}, \hat{a}_{d}) = R_c(a_{c}, a_{d}, Y) - R_c(\hat{a}_{c}, \hat{a}_{d}, Y)
\end{equation}
and $\hat{a}_{c}$ and $\hat{a}_{d}$ represent the baseline action vectors. To get $\hat{a}_{c}$, we use the most likely action vector proposed by the policy network: \textit{i.e.}, $a_{c}^{i}=1$ if $s_{c}^{i}>0.5$ and $a_{c}^{i}=0$ otherwise. The coarse and fine level detectors, $f_{d}^{c}$ and $f_d^f$, then observes $x_{L}^{i}$ or $x_{H}^{i}$, and outputs the predicted bounding boxes $\hat{a}_{d}$. The advantage function assigns the policy network \emph{a positive value} only when the action vector sampled from Eq.~\ref{eq:action_likelihood} produces higher reward than the action vector with maximum likelihood, which is known as a self-critical baseline~\cite{rennie2017self}. Similarly to the coarse level search, we introduce the advantage function to the \emph{fine level search} module and do not show it in this section for simplicity.

Finally, in this study we use the temperature scaling~\cite{sutton2018reinforcement} to encourage exploration during training time by bounding the probabilities of the policy networks as
\begin{equation}
s = \alpha s + (1-\alpha)(1-s).
\end{equation}
In our experiments, we tune $\alpha$ to $0<\alpha<1$, where we observe that setting it to a small value produces more uniform probabilities to sample off-policy actions.

\subsection{Modeling the Reward Function}
\label{sect:reward_function}
The proposed framework uses the policy gradient reinforcement learning algorithm to learn the parameters of the policy networks, adjusting their weights to increase the expected reward value. Thus, it is crucial to design a reward function reflecting the desired characteristics of an efficient object detection method for large images: low \emph{image acquisition cost} and high \emph{run-time efficiency}. For the coarse level policy network, our reward function $R_{c}$ encourages the use of LR image patches $x_{L}^{i}$ with the coarse level detector $f_{d}^{c}$. We define $R_c$ as follows, where $\hat{Y}^f$ are the object detections by the fine-level object detector on patches of $x_H$, and $\hat{Y}^c$ are the detections by the coarse-level detector on $x_L$:
\begin{equation}
    R_c = R_{acc}(\hat{Y}^{f}, \hat{Y}^{c}, Y) + R_{cost}(a_{c})
\end{equation}
\begin{equation}
    R_{acc} = \sum_{i=1}^{P_{c}}(Recall(\hat{Y}^{f}_{i}, Y_{i}) - (Recall(\hat{Y}^{c}_{i}, Y_{i})+\beta)) \cdot N_{i} \label{eq:reward_function_acc}
\end{equation}
\begin{equation}
    R_{cost} = (\sigma + \lambda)(1 - |a_{c}|_1) / P_{c}
\end{equation}
where $R_{acc}$ is detection recall and $R_{cost}$ combines image acquisition cost and run-time performance reward. The $R_{acc}$ term encourages zooming-in when the recall difference between the coarse and fine detector is positive. The difference is then scaled with the number of objects in the patch, $N_{i}$, to prioritize zooming-in to regions where \emph{there are more objects}. $\beta$, on the other hand, prioritizes using LR images when the recall values from coarse and fine detector are similar. Note that since our priority is to \emph{minimize the ratio of false negatives}, we only use Recall and do not consider \emph{Precision} in $R_{acc}$. The other component of reward, $R_{cost}$, represents the run-time and image acquisition costs with their own coefficients $\lambda$ and $\sigma$. In this case, the reward increases \emph{linearly} with the smaller number of zoomed-in patches. The reward function $R_f$ for the fine-level policy network is defined similarly.

\section{Experiments}
\subsection{Baselines and State-Of-The-Art Models}
\textbf{Sliding Window}
A simple method for running object detectors on large images is the \emph{sliding window} approach. Using this fixed policy approach, we either apply a coarse level detector, $f_{d}^{c}$, on $x_{l}^j$ or a fine level detector, $f_{d}^{f}$, on $x_{h}^{j}$ for $j=1, \dotsc, P_{f}$. We then repeat this for every $x_{H}^{i}$ or $x_{L}^{i}$ for $i=1, \dotsc, P_{c}$. 

\textbf{Random Policy}
A simple adaptive policy can be designed with a random policy. In this case, we can sample patch specific probabilities from a policy represented by patch specific uniform distribution.


\textbf{Entropy Based Policy Sampling}
A more sophisticated policy can be developed using an entropy based approach. In this case, we first the compute the patch-wise confidence of the coarse level object detector, $s_{c}^{i}=\frac{1}{M} \sum_{m=1}^M c_m$, on $x_{L}^{i}$ where $M$ represents the number of detected bounding boxes. Next, we pass the patch to the fine level search if $s_{c}^{i}$ is larger than a threshold. In the fine level search, we compute $s_{f}^{j}$ similarly to the coarse level search given that $a_{c}^{i}=1$ and acquire a HR subpatch $x_{h}^{j}$ if $s_{f}^{j}$ is larger than a threshold and use it for fine level object detection. Otherwise, we use $x_{l}^{j}$ for coarse level object detection. 


\textbf{Dynamic Zoom-in Network} \cite{gao2018dynamic} proposed the current \emph{state-of-the-art} method for efficiently detecting objects in large images without changing the underlying structure of the detector. They show the results on \emph{640x480} pixels images from the Caltech Pedestrian Detection dataset~\cite{dollar2009pedestrian}. Their method is not suitable for larger images, i.e. $\sim$3000 pixels, as it starts with a detector trained on \emph{full images} downsampled by 2. To make it practical for larger images, we train their coarse detector on the full images downsampled by 5 similar to our coarse detector. Since their code is not available publicly, we implemented it in PyTorch.

\textbf{Variants of the Proposed Approach}
Additionally, in this section, we report the \emph{coarse level only} and \emph{fine level only} results for both the baseline models and proposed approach. In the coarse level only method, we run the coarse level policy network on $x_{L}$ and acquire HR image $x_{H}^{i}$ if $a_{c}^{i}$=1 and use it for fine level object detection. In the fine level only approach, we run the fine level policy network on $x_{L}^{i}$ unconditioned to $a_c$. Then, we acquire a HR image $x_{h}^{j}$ if $a_{f}^{j}$=1 and use it for fine level object detection. 
\subsection{Implementation Details}
\textbf{Policy Networks} To parameterize the coarse and fine level policy networks, we use ResNet \cite{he2016deep} with 32 layers pretrained on the ImageNet Large Scale Visual Recognition Challenge 2012 (ILSVRC2012) dataset~\cite{russakovsky2015ilsvrc}. We train the policy networks using 4 NVIDIA 1080ti GPUs.

\textbf{Object Detectors} Our coarse and fine level detectors use the YOLOv3 architecture~\cite{redmon2018yolov3}, chosen for its reasonable trade off between accuracy on small objects and run-time performance. The backbone network, DarkNet-53, is pre-trained on ImageNet. We train the detectors using a single NVIDIA 1080ti GPU. Our code for this work can be found at the following repository\footnote{\color{blue}{https://github.com/uzkent/EfficientObjectDetection}}.

\subsection{Performance Metrics}
We evaluate the performance using the following metrics: \emph{average precision (AP)}, \emph{average recall (AR)}, \emph{average run-time per image (ms)}, and \emph{ratio of sampled HR image}. For AP and AR, we compute the individual values across different categories for IoU=$\{.50,.55,.60,\dotsc,.95\}$.

\begin{table*}[ht]
\centering
\resizebox{\textwidth}{!}{
\begin{tabular}{@{}lclllclllclll@{}}
\toprule
 & \multicolumn{4}{c}{Coarse Level} & \multicolumn{4}{c}{Fine Level} & \multicolumn{4}{c}{Coarse + Fine Level} \\ \midrule
Model/Metric & \multicolumn{1}{l}{AP} & \multicolumn{1}{l}{AR} & Run-time & HR & \multicolumn{1}{l}{AP} & \multicolumn{1}{l}{AR} & Run-time & HR &  \multicolumn{1}{l}{AP} &  \multicolumn{1}{l}{AR} & Run-time & HR \\
Random (5$\times$) & \multicolumn{1}{l}{29.2} & 47.0 & 1770 & 43.7 & \multicolumn{1}{l}{27.2} & 49.3 & 1920 & 50 & \multicolumn{1}{l}{24.1} & 47.1 & 1408 & 31\\
Entropy (5$\times$) & \multicolumn{1}{l}{30.1} & 47.9 & 1766 & 43.7 & \multicolumn{1}{l}{28.3} & 50.1 & 1932 & 50 & \multicolumn{1}{l}{25.4} & 47.2 & 1415 & 31\\
Sliding Window-L (5$\times$) & \multicolumn{1}{l}{26.3} & 39.8 & 640 & 0 & \multicolumn{1}{l}{26.3} & 39.8 & 640 & 0 & \multicolumn{1}{l}{26.3} & 39.8 & 640 & 0 \\
Sliding Window-H &  \multicolumn{1}{l}{39.0} & 60.9 & 3200 & 100 & \multicolumn{1}{l}{39.0} & 60.9 & 3200 & 100 & \multicolumn{1}{l}{39.0} & 60.9 & 3200 & 100 \\
Gao et al.~\cite{gao2018dynamic} (5$\times$) &  \multicolumn{1}{l}{35.3} & 55.2 & 1780 & 40.5 & \multicolumn{1}{l}{35.2} & 55.8 & 1721 & 35.4 & \multicolumn{1}{l}{35.2} & 55.5 & 1551 & 31.6\\
\textbf{Ours} (5$\times$) & \multicolumn{1}{l}{38.2} & 59.8 & 1725 & 40.6 & \multicolumn{1}{l}{38.3} & 59.6 & 1683 & 35.5 & \multicolumn{1}{l}{38.1} & 59.7 & 1484 & 31.5 \\ \bottomrule
\end{tabular}}
\caption{Results for the \textit{building} and \textit{small car} classes. The coarse and fine level only methods refer to using only coarse and fine level policy network in test time. The coarse and fine level method first runs the coarse level policy network on initial large image, and fine level policy network is run on the images activated by the coarse network.} \label{tab:building_results}
\end{table*}
\begin{figure*}[!ht]
\centering
\includegraphics[width=0.98\textwidth]{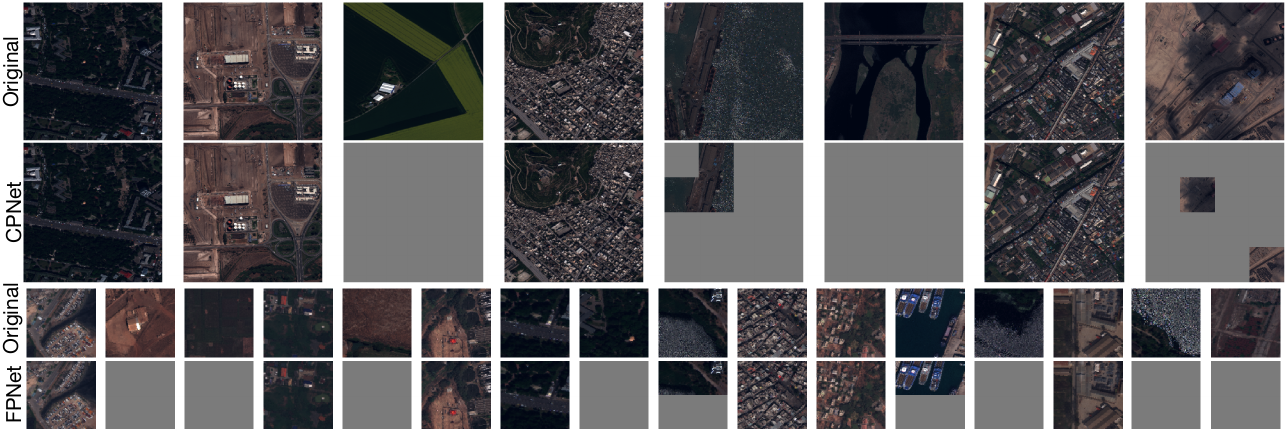}
\caption{Visualization of the learned policies by the coarse and fine level policy networks. The top row shows the original large LR images given to the coarse policy network. The second row shows the patches chosen by the coarse policy network, and last two rows represent the policy learned by the fine level policy network. The coarse and fine level policy networks learn to sample LR-HR patches/sub-patches, respectively. The regions for using LR images are shown in grey.} \label{fig:visualize_policy_building}
\vspace{-0.5em}
\end{figure*}
\subsection{Experiments on xView Dataset}
We evaluate the proposed approach and baseline methods on the xView dataset~\cite{lam2018xview}, which consists of large HR satellite images representing 60 categories. The training, validation, and test splits of the dataset have 846, 221, and 221 large scale images, respectively, with 3000-6000 pixels in each dimension. The validation and test splits of the dataset have not been released publicly since the xView dataset was collected as part of the Object Detection challenge on Satellite Images. For this reason, we use the training split of the dataset to train the object detectors and policy networks and test the proposed framework. In particular, $47\%$, and $12\%$ of the large images is used to train and test the coarse and fine level detector whereas the remaining $41\%$ is used to train the policy network. The policy network is then tested on the same $12\%$ of the large images used to test the detectors. Among the 60 classes in the xView dataset, we test our efficient object detection model on the \emph{small car} and \emph{building} classes as they are the two classes represented by the most number of samples.

\textbf{Coarse and Fine Level Detectors}
In the first step, we train the coarse and fine level detectors. We train the coarse and fine level detectors with the batch size of 8 for 65 and 87 epochs, respectively. The coarse and fine level detectors achieve $26.3\%$ and $39.8\%$ average precision (AP) and $39.0\%$ and $60.9\%$ average recall (AR). The coarse detector operates on the 64$\times$64px LR images, $x_l^j$, whereas the fine level detector operates on the 320$\times$320px HR images, $x_h^j$. In other words, 320$\times$320px images, $x_{h}^{j}$, correspond to a subpatch of 600$\times$600px images, $x_{H}^{i}$, where we set the overlap between subpatches to 40 pixels. The coarse and fine detectors run on average at $10$ and $50$ ms per image on a NVIDIA GeForce GTX 1080 Ti GPU.

\textbf{Policy Networks} In the second step, we train the policy networks and treat the detectors as \emph{black boxes} as we keep their weights fixed. We form the initial large images by taking 2400$\times$2400px crops out of the xView images, then divide them into 16 non-overlapping 600$\times$600px patches resulting in 16 output unit in CPNet (See Appendix). For the CPNet, we downsample the images by a factor of about 5 to 448$\times$448px. For the FPNet, we use 4 output units to represent the subpatches in the patch selected by the CPNet (See Appendix). The subpatches have the size and overlap of 320$\times$320px and 40 pixels. As the input, we use the 112$\times$112px patches of the 448$\times$448px images used by CPNet. 
Then, we train CPNet with a batch size of 512 for 643 epochs. In the final step, we train FPNet with a batch size of 512 for 459 epochs. For both networks, we set the learning rate to 1e-4 and the hyperparameters $\alpha$, $\lambda$, $\sigma$ and $\beta$ to 0.8, 0.25, 0.25 and 0.05. All the networks are trained with the Adam optimizer~\cite{kingma2015adam}. See \textbf{Appendix} for visualization of action space for CPNet and FPNet.

\textbf{Quantitative and Qualitative Analysis} As shown in Table~\ref{tab:building_results}, the cascaded approach using CPNet+FPNet provides optimal results considering the run-time efficiency and image acquisition cost. It yields AP and AR scores only $0.9\%$ and $1.2\%$ lower, respectively, than the sliding window approach using HR images with the fine detector. Meanwhile, our approach uses HR images for the $31.5\%$ of the full area of interest on average and delivers 2.2$\times$ higher run-time efficiency. On the other hand, the CPNet+FPNet approach in test time outperforms CPNet and FPNet only approach in terms of the use of HR image and run-time efficiency. Additionally, Fig.~\ref{fig:visualize_policy_building} demonstrates the learned policies by CPNet and FPNet. For instance, in the columns 1, 2 and 4, CPNet learns to use HR images as these images mostly contain small buildings and cars. On the other hand, it learns to use LR images when the image is populated with no objects or large buildings (columns 3 and 8). Similarly, FPNet chooses HR images when the subpatches are populated with small buildings or cars as in columns 1, 4, 10 and 11.
\begin{table*}[!h]
\centering
\resizebox{\textwidth}{!}{
\begin{tabular}{@{}lclllclllclll@{}}
\toprule
 & \multicolumn{4}{c}{Coarse Level} & \multicolumn{4}{c}{Fine Level} & \multicolumn{4}{c}{Coarse + Fine Level} \\ \midrule
Model/Metric & \multicolumn{1}{l}{AP} & \multicolumn{1}{l}{AR} & Run-time & HR & \multicolumn{1}{l}{AP} & \multicolumn{1}{l}{AR} & Run-time & HR & \multicolumn{1}{l}{AP} & \multicolumn{1}{l}{AR} & Run-time & HR \\
Random (5$\times$) & \multicolumn{1}{l}{22.7} & 32.4 & 1792 & 56.1 & \multicolumn{1}{l}{21.5} & 31.4 & 1601 & 50.0 & \multicolumn{1}{l}{19.9} & 29.8 & 1504 & 47.1 \\
Entropy (5$\times$) & \multicolumn{1}{l}{23.0} & 32.2 & 1801 & 56.2 & \multicolumn{1}{l}{22.2} & 30.6 & 1605 & 50.0 & \multicolumn{1}{l}{20.5} & 31.1 & 1511 & 47.5 \\
Sliding Window-H & \multicolumn{1}{l}{39.0} & 60.9 & 3200 & 100 & \multicolumn{1}{l}{39.0} & 60.9 & 3200 & 100 & \multicolumn{1}{l}{39.0} & 60.9 & 3200 & 100 \\
\textbf{Ours} (5$\times$) & \multicolumn{1}{l}{37.4} & 58.1 & 1882 & 58 & \multicolumn{1}{l}{37.5} & 58.2 & 1640 & 45.1 & \multicolumn{1}{l}{37.3} & 58.1 & 1421 & 44.1\\ \bottomrule
\end{tabular}}
\caption{Results for the \textit{building} and \textit{small car} classes when the coarse detector is \emph{disabled}. The coarse detector is removed in the training and test time. We removed the method by Gao et al.~\cite{gao2018dynamic} as its agent uses coarse detector to zoom-in.}
\label{tab:remove_coarse_detector}
\vspace{-0.21cm}
\end{table*}

\begin{figure}[!b]
\vspace{-1.0em}
\includegraphics[width=0.235\textwidth]{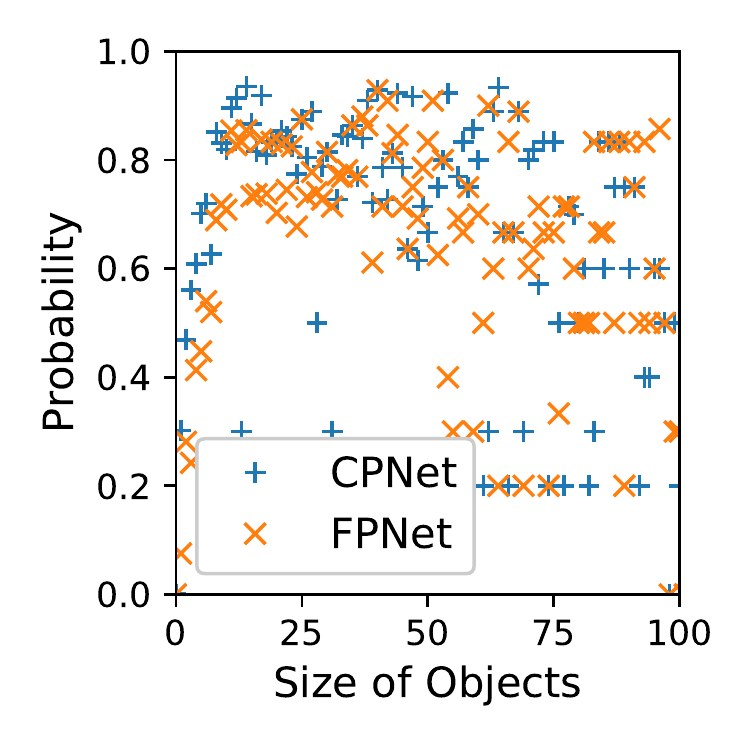}
\includegraphics[width=0.235\textwidth]{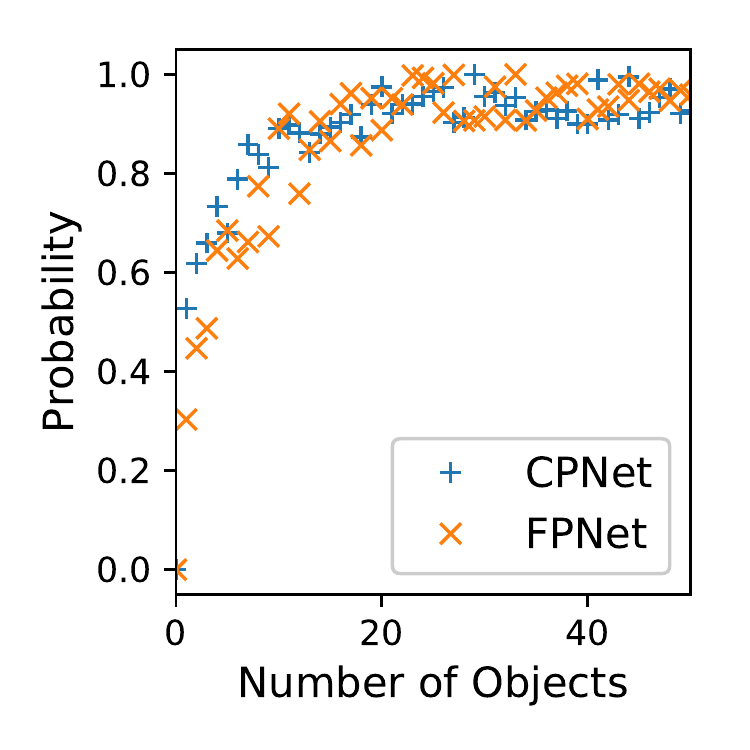}
\caption{Visualization of when the CPNet and FPNet zooms-in w.r.t. the average number of objects and size of objects. Size corresponds to the normalized area of the objects in pixels.
}
\label{fig:objects_vs_probability}
\end{figure}

Finally, in Fig.~\ref{fig:objects_vs_probability} we show the probability of zoom-in for the CPNet and FPNet when the number and size of objects in each patch/subpatch increase. The probability of zoom-in increases w.r.t. number of objects because we scale $R_{acc}$ with the number of objects $N_i$ in a patch/subpatch. On the other hand, the increasing average size of the objects does not necessarily mean an increased probability of zoom-in. This can be expected as the recall difference between the coarse and fine detector reduces with increasing size of the objects.


\textbf{Ablation Experiments on xView}
Previously, we already performed ablation experiments (Table~\ref{tab:building_results}) by excluding coarse (CPNet) or fine level search (FPNet) from our cascaded approach. 
In this section, we want to quantify the effect of the coarse detector on CPNet, FPNet and the baseline methods. By removing the coarse level detector from Eq.~\ref{eq:reward_function_acc}, we can understand if the policy networks can learn the difference between \emph{zooming-in when there is object of interest} and \emph{zooming-in when the fine level detector outperforms coarse level detector}. Thus, we train the CPNet and FPNet using the modified reward function. In particular, we encourage the network to zoom-in when the fine detector achieves a positive recall value on HR image. 

Table~\ref{tab:remove_coarse_detector} shows that removing the coarse detector improves the run-time efficiency slightly with the cost of using about $13\%$ more HR images.
This can be justified with the fact that maintaining the accuracy requires zooming-in to more patches with the removal of coarse detector. In conclusion, by integrating the coarse detector we can achieve better accuracy and similar run-time performance while relying on fewer HR images.

\subsection{Experiments on Caltech Pedestrian Dataset}
Finally, we run experiments on the Caltech Pedestrian dataset (CPD)~\cite{dollar2009pedestrian} to quantify the validity of the proposed approach on \emph{traditional images}. We resize the original $640\times480$px images to $860\times860$px similarly to~\cite{gao2018dynamic}. Following \emph{reasonable} setting in CPD, we use two sets of 5000 images for training the detectors and policy networks, and two sets of 3000 images for validation and test experiments.

\textbf{Coarse and Fine Level Detectors} We first train the coarse and fine detectors. The fine detector is trained on $320\times320$px patches of the $860\times860$px images and coarse detector is trained on downsampled version of the patches. 

\textbf{Policy Network} Since the images are $\sim 3\times$ smaller than xView images, we only use CPNet and form its action space with \emph{9 unique patches}. In this setting, each patch has size $320\times320$px, and the overlap between the patches is $50$ pixels (See Appendix). For CPNet, we use $172\times172$px images downsampled from $860\times860$px to maintain the same $5 \times$ downsampling ratio as in our xView experiments. We set the other parameters similarly to our xView experiments and train the CPNet for 1450 epochs. See \textbf{Appendix} for visualization of action space for CPNet.

\textbf{Quantitative Analysis} As shown in Table~\ref{tab:cpd}, the proposed approach outperforms the baselines and state-of-the-art~\cite{gao2018dynamic} by a large margin when using a $5\times$ downsampling ratio. When using $2\times$ downsampling, our method performs similarly to~\cite{gao2018dynamic} in terms of AP and AR, but delivers higher run-time performance. This is because their agent uses coarse detection results over the entire image, whereas CPNet only chooses a binary action for each image patch without using coarse detection, thus requiring less overhead.
Our approach can be even further optimized by running the patches through the coarse or fine detectors \emph{in parallel}, whereas~\cite{gao2018dynamic} performs \emph{sequential} inference.
\begin{table}[!h]
\centering
\resizebox{0.46\textwidth}{!}{
\begin{tabular}{@{}lllll@{}}
\toprule
Model/Metric & AP & AR & Run-time & HR \\ \midrule
Random ($\times$5) & 30.9 & 62.1 & 248 & 44.4 \\
Entropy ($\times$5)  & 34.0 & 63.9 & 250 & 44.4 \\
Sliding Window-L ($\times$5) & 21.2 & 46.3 & 90 & 0 \\
Sliding Window-H & 64.7 & 74.7 & 450 & 100 \\
Gao et al.~\cite{gao2018dynamic} ($\times$2) & 64.5 & 73.1 & 295 & 7.1 \\
Gao et al.~\cite{gao2018dynamic} ($\times$5) & 57.3 & 70.7 & 309 & 43.3 \\
\textbf{CPNet} ($\times$2) & 64.4 & 74.5 & 267 & 6.6 \\
\textbf{CPNet} ($\times$5) & 61.7 & 74.1 & 270 & 44.4 \\ \bottomrule
\end{tabular}}
\caption{Results on the Caltech Pedestrian Dataset. We show the visuals representing the policies learned by CPNet in \textbf{Appendix}.}
\label{tab:cpd}
\end{table}
\begin{figure}[!h]
\vspace{-1.0em}
\includegraphics[width=0.23\textwidth]{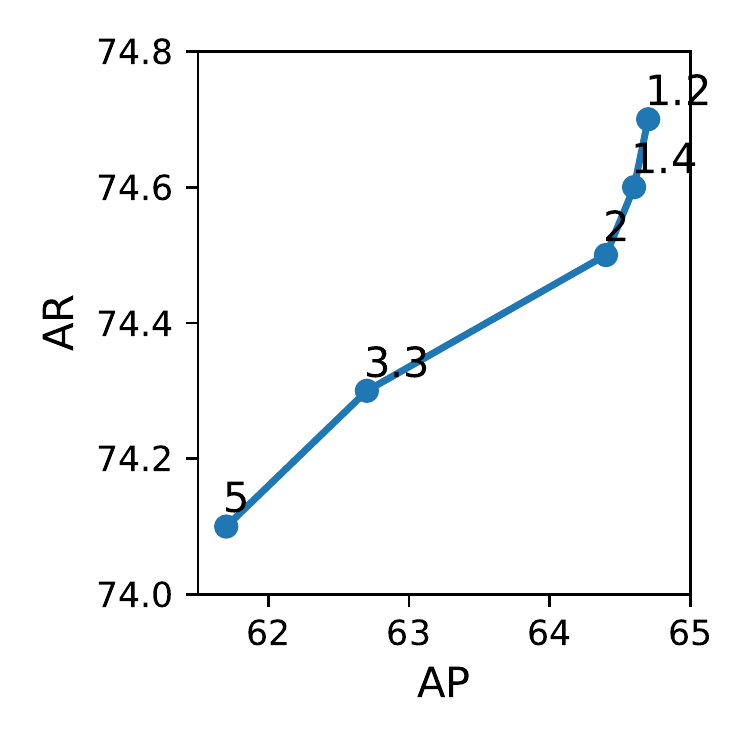}
\includegraphics[width=0.23\textwidth]{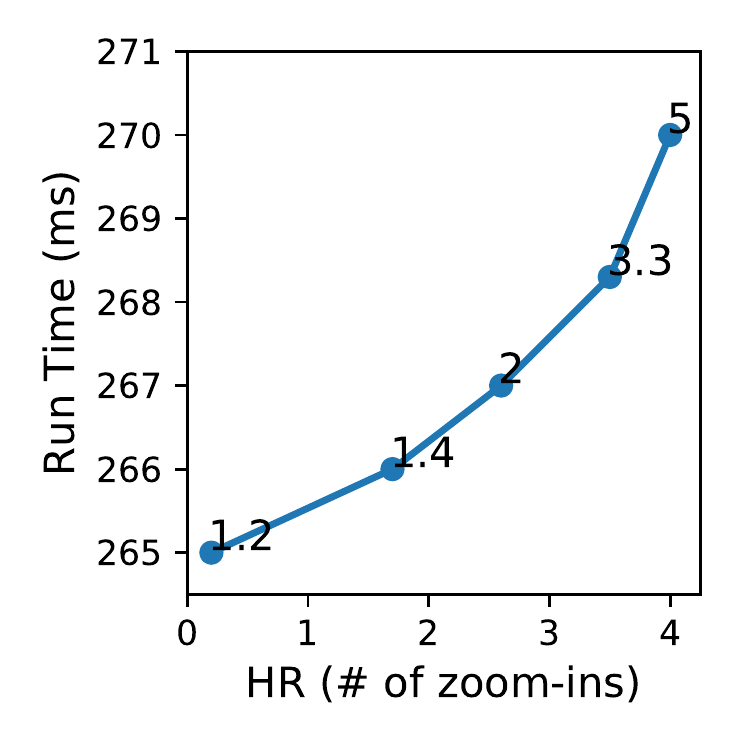}
\caption{Performance of the CPNet when changing the resolution of the coarse detector on CPD. We use 172$\times$172px images for the CPNet and only change the resolution of coarse detector.}
\label{fig:vs_downsampling}
\end{figure} 

Finally, we measure the performance of the CPNet w.r.t. different downsampling factors for the coarse detector. As seen in Fig.~\ref{fig:vs_downsampling}, the AP, AR and Run-time do not change drastically with decreasing downsampling ratio whereas the number of zoom-ins decreases sharply, since the coarse detector performs better with finer images but also costs larger run-time. These results show that CPNet can maintain high AP, AR and run-time efficiency when used with the coarse detector trained on images across different resolutions.

\section{Conclusion}
In this study, we proposed an approach to efficiently process large images for object detection without changing the underlying structure of a detector. In particular, we trained two policy networks, CPNet and FPNet, using reinforcement learning with the dual reward of maintaining the accuracy while maximizing the use of LR images with a coarse detector. By choosing actions for the full image in one step, the policy networks introduce minimal overhead. Our experiments on the xView, consisting of very large satellite images, indicate that the proposed approach increases run-time efficiency by 2.2$\times$ while reducing dependency on HR images by about $70\%$. Finally, our approach delivers $40\%$ run-time increase on Caltech Pedestrian Dataset images.

\section*{Acknowledgements}
This research was supported by Stanford's Data for Development Initiative and NSF grants 1651565 and 1733686.

{\small
\bibliographystyle{ieee}
\bibliography{egbib}
}

\newpage
\section*{Appendix}
\begin{figure*}[!t]
\centering
\includegraphics[width=0.98\textwidth]{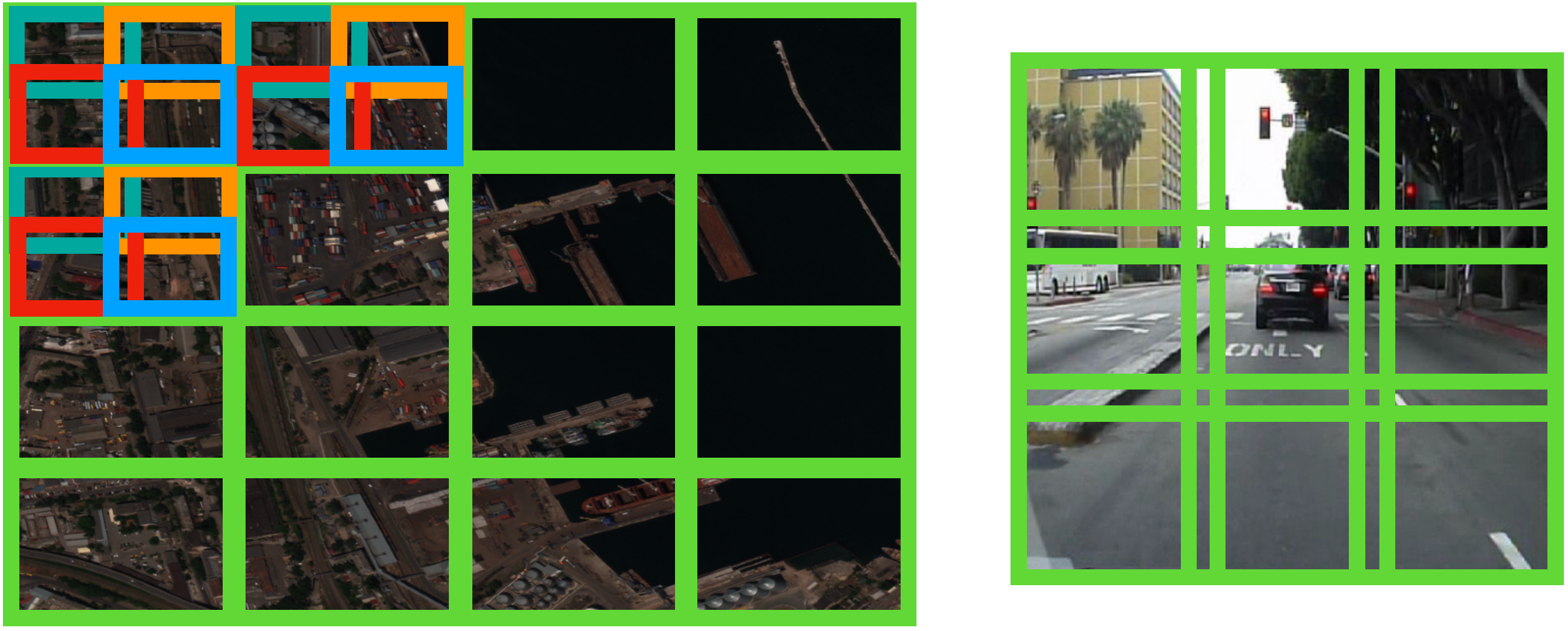}
\caption{Visualization of the action spaces for the xView (\textbf{Left}) and CPD (\textbf{Right}) experiments. Green patches represent the patches for the CPNet whereas the differently colored subpatches represent the subpatches for the FPNet. For simplicity, we show the FPNet subpatches only on the 3 patches. The size of the patches for the xView images is 600$\times$600px whereas the subpatches are 320$\times$320px. The size of the patches for the CPD images is 320$\times$320px. The fine detector for the xView is trained on 320$\times$320px subpatches whereas for the CPD it is trained on 320$\times$320px patches. The coarse detectors, on the other hand, are trained on downsampled version of these images.} 
\end{figure*}

\begin{figure*}[!t]
\centering
\includegraphics[width=0.98\textwidth]{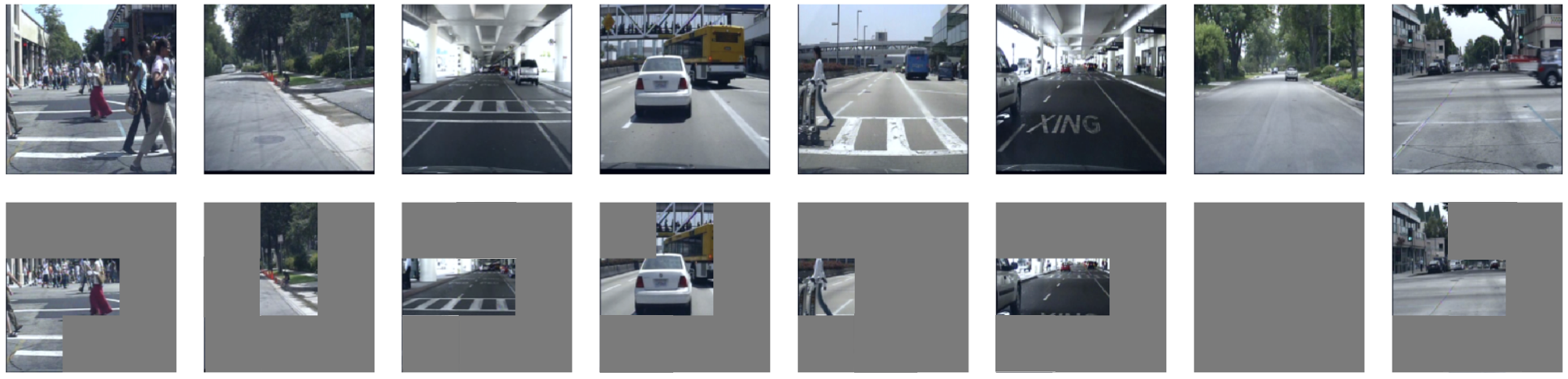}
\caption{Visualization of the learned policies by the CPNet on Caltech Pedestrian Dataset. The CPNet zooms-in when there is pedestrians represented with relatively small number of pixels in the scene as seen in columns 1, 3, and 4. On the other hand, it also zooms in to ROIs where there might be pedestrians represented with very small number of pixels as in columns 2, 6, and 8. The regions for using LR images are shown in grey.} 
\end{figure*}

\end{document}